\documentclass[runningheads]{llncs}
\usepackage{graphicx, picinpar,wrapfig}
\usepackage{graphicx}
\usepackage{amsmath,amssymb} 
\usepackage{multirow}
\usepackage{color,subfigure}
\usepackage{capt-of}
\graphicspath{{/}}
\DeclareGraphicsExtensions{.pdf,.jpeg,.png,.eps,.jpg,.jpeg}
\DeclareMathOperator*{\argmaxA}{arg\,max} 
\DeclareMathOperator*{\argminA}{arg\,min}
\begin{document}

\title{Stereo Vision-based Semantic 3D Object and Ego-motion Tracking for Autonomous Driving} 

\titlerunning{Semantic 3D Object and Ego-motion Tracking}

\authorrunning{P. Li, T. Qin, and S. Shen}

\author{Peiliang Li, Tong Qin, and Shaojie Shen}
	

\institute{Hong Kong University of Science and Technology\\
	\email{ \{pliap,tong.qin,eeshaojie\}@ust.hk}
}

\maketitle
\begin{abstract}
We propose a stereo vision-based approach for tracking the camera ego-motion and 3D semantic objects in dynamic autonomous driving scenarios. Instead of directly regressing the 3D bounding box using end-to-end approaches, we propose to use the easy-to-labeled 2D detection and discrete viewpoint classification together with a light-weight semantic inference method to obtain rough 3D object measurements. 
Based on the object-aware-aided camera pose tracking which is robust in dynamic environments, in combination with our novel dynamic object bundle adjustment (BA) approach to fuse temporal sparse feature correspondences and the semantic 3D measurement model, we obtain 3D object pose, velocity and anchored dynamic point cloud estimation with instance accuracy and temporal consistency. The performance of our proposed method is demonstrated in diverse scenarios. Both the ego-motion estimation and object localization are compared with the state-of-of-the-art solutions.
\end{abstract}

\section{Introduction}

Localizing dynamic objects and estimating the camera ego-motion in 3D space are crucial tasks for autonomous driving.  Currently, these objectives are separately explored by end-to-end 3D object detection methods \cite{chen20153d,chen2016monocular} and traditional visual SLAM approaches \cite{qin2017vins,murORB2,engel2014lsd}. However, it is hard to directly employ these approaches for autonomous driving scenarios.
For 3D object detection, there are two main problems: 1. end-to-end 3D regression approaches need lots of training data and require heavy workload to precisely label all the object boxes in 3D space and 2. the instance 3D detection produces frame-independent results, which are not consistent enough for continuous perception in autonomous driving. 
To overcome this, we propose a light-weight semantic 3D box inference method depending only on 2D object detection and discrete viewpoint classification (Sect. \ref{sec:viewpoint}). Comparing with directly 3D regression, the 2D detection and classification task are easy to train, and the training data can be easily labeled with only 2D images. However, the proposed 3D box inference is also frame-independent and conditional on the instance 2D detection accuracy.
\begin{figure}
	\setlength{\belowcaptionskip}{-0.5cm} 
	\centering
	\includegraphics[width=0.97\columnwidth]{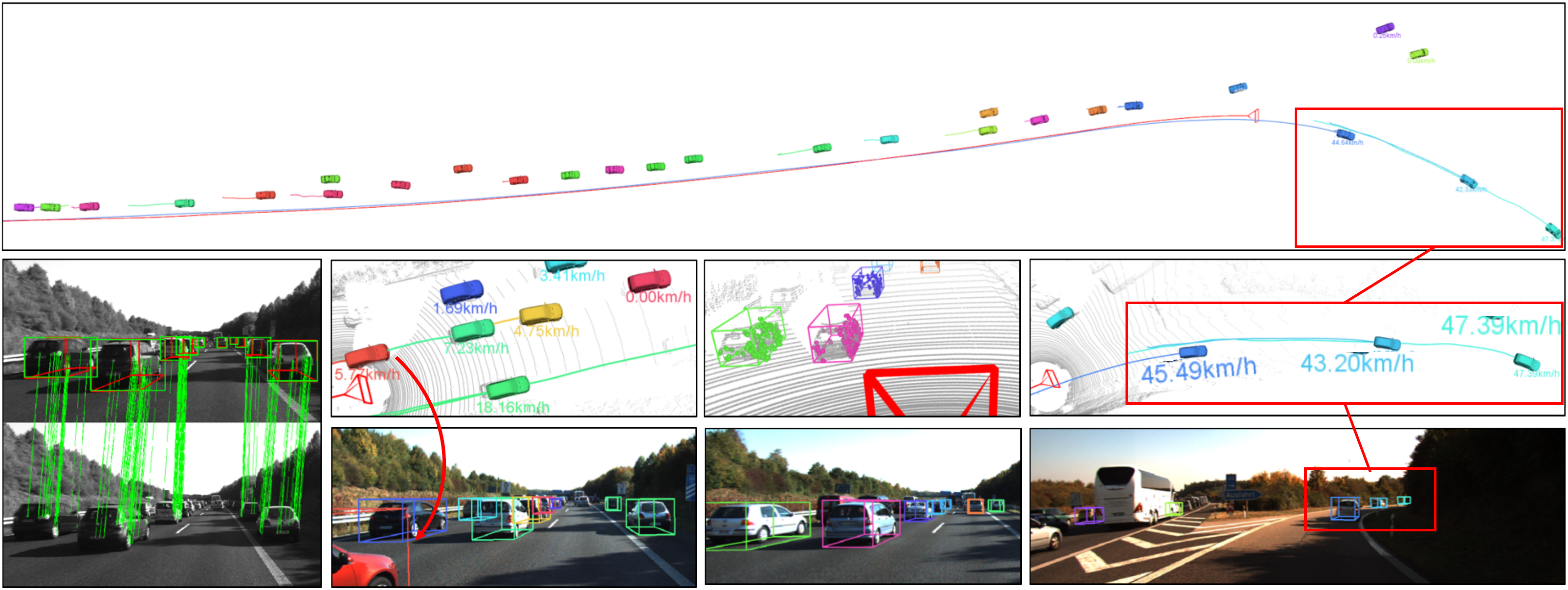}
	\caption{Overview of our semantic 3D object and ego-motion tracking system. Top: 3D trajectories of ego-camera and all objects in the long travel history. Bottom: From left to right: Stereo feature matching for each object (Sect. \ref{sec:matching}). An extreme car-truncated case where our system can still track the moving car accurately. Dynamic 3D sparse feature recovered by our object BA. Consistent movement and orientation estimation.
		\label{fig:feature}}
\end{figure}
In another aspect, the well-known SLAM approaches can track the camera motion accurately due to precise feature geometry constraints.
Inspired by this, we can similarly utilize the sparse feature correspondences for object relative motion constraining to enforce temporal consistency. However, the object instance pose cannot be obtained with pure feature measurement without semantic prior.
To this end, due to the complementary nature of
semantic and feature information, we integrate our instance semantic inference model and the temporal feature correlation into a tightly-coupled optimization framework which can continuously track the 3D objects and recover the dynamic sparse point cloud with instance accuracy and temporal consistency, which can be overviewed in Fig. \ref{fig:feature}. 
Benifitting from object-region-aware property, our system is able to estimate camera pose robustly without being affected by dynamic objects.
Thanks to the temporal geometry constraints, we can track the objects continuously even for the extremely truncated case (see Fig. \ref{fig:feature}), where the object pose is hard for instance inference. Additionally, we employ a kinematics motion model for detected cars to ensure consistent orientation and motion estimation; it also serves as important smoothing for distant cars which have few feature observation. Depending only on a mature 2D detection and classification network \cite{ren2015faster}, our system is capable of performing robust ego-motion estimation and 3D object tracking in diverse scenarios. The main contributions are summarized as follows:
\begin{itemize}
	\item A light-weight 3D box inference method using only 2D object detection and the proposed viewpoint classification, which provides the object reprojection contour and occlusion mask for object feature extraction. It also serves as the semantic measurement model for the follow-up optimization.
	\item A novel dynamic object bundle adjustment approach which tightly couples the semantic and feature measurements to continuously track the object states with instance accuracy and temporal consistency.
	\item Demonstration over diverse scenarios to show the practicability of the proposed system.
\end{itemize}

\section{Related Work}

We review the related works in the context of semantic SLAM and learning-based 3D object detection from images.

\subsubsection{Semantic SLAM}

With the development of 2D object detection, several joint SLAM with semantic understanding works have sprung up, which we discuss in three categories. 
The first is semantic-aided localization: \cite{frost2016object,DBLP:journals/corr/SucarH17} focus on correcting the global scale of monocular Visual Odometry (VO) by incorporating object metric size of only one dimension into the estimation framework. Indoor with small objects and outdoor experiments are conducted respectively in these two works. \cite{bowman2017probabilistic} proposes an object data association method in a probabilistic formulation and shows its drift correcting ability when re-observing the previous objects. However, it omits the orientation of objects by treating the 2D bounding boxes as points. And in \cite{atanasov2014semantic}, the authors address the localization task from only object observation in a prior semantic map by computing a matrix permanent. 
The second is SLAM-aided object detection \cite{recognition,DBLP:journals/corr/Soatto16} and reconstruction \cite{civera2011towards,salas2013slam++}: \cite{recognition} develops an 2D object recognition system which is robust to viewpoint changing with the assistance of camera localization, while \cite{DBLP:journals/corr/Soatto16} performs confidence-growing 3D objects detection using visual-inertial measurements. \cite{civera2011towards,salas2013slam++} reconstruct the dense surface of 3D object by fusing the point cloud from monocular and RGBD SLAM respectively.
Finally, the third category is joint estimation for both camera and object poses: With pre-built bags of binary words, \cite{galvez2016real} localizes the objects in the datasets and correct the map scale in turns. In \cite{bao2011semantic,bao2012semantic}, the authors propose a semantic structure from motion (SfM) approach to jointly estimate camera, object with considering scene components interaction. However, neither of these methods shows the ability to solve dynamic objects, nor makes full use of 2D bounding box data (center, width, and height) and 3-dimensions object size. There are also some existing works \cite{kundu2014joint,vineet2015incremental,li2016semi,mccormac2017semanticfusion} building the dense map and segmenting it with semantic labels. These works are beyond the scope of this paper, so we will not discuss them in details.

\begin{figure}
	\centering
	\includegraphics[width=1.0\columnwidth]{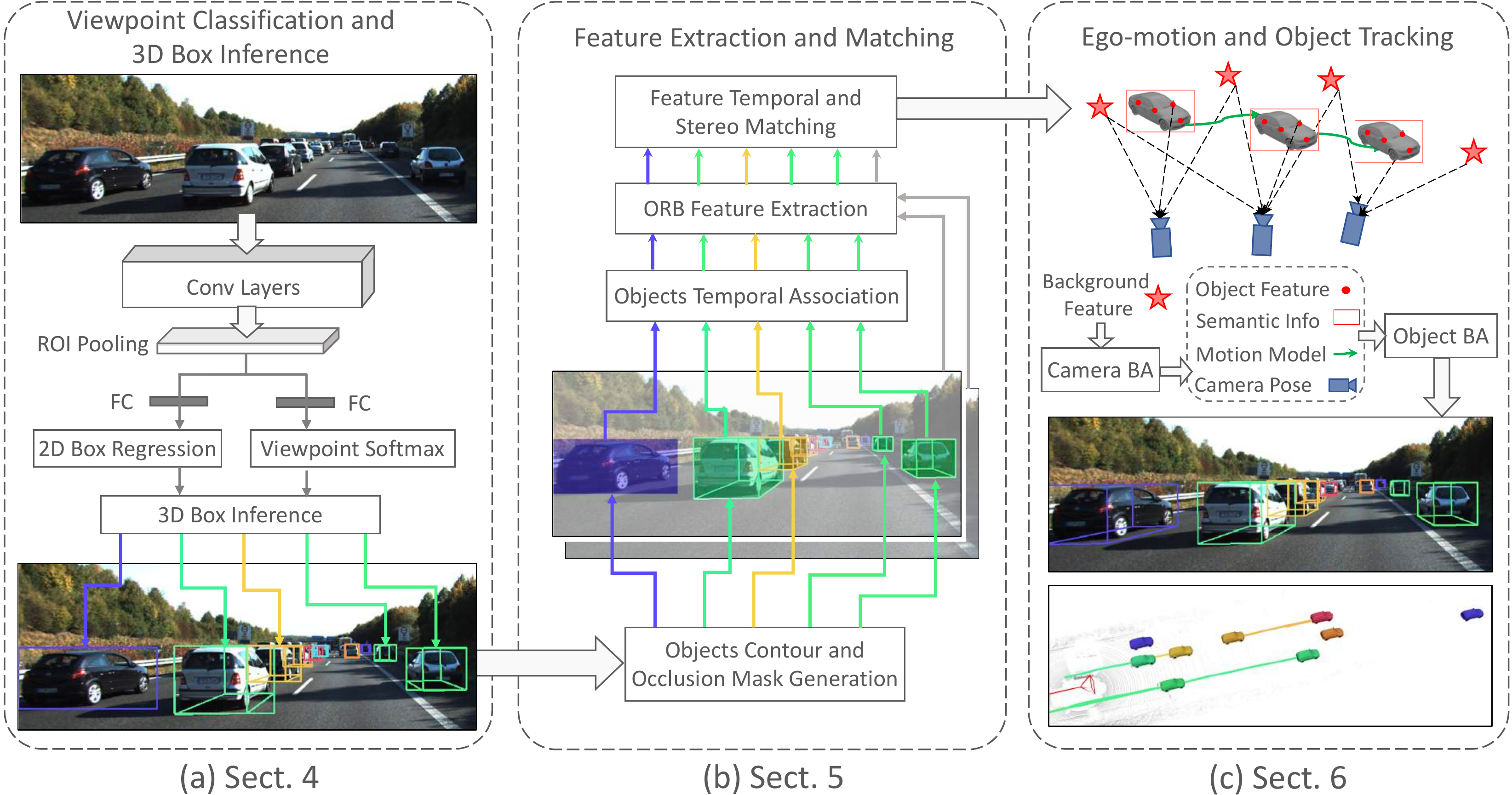}
	\caption{Our whole semantic tracking system architecture.
		\label{fig:system}}
\end{figure}

\subsubsection{3D Object Detection} Inferring object pose from images by deep learning approaches provides an alternative way to localize 3D objects. \cite{bao2013dense,zia2013detailed} use the shape prior to reason 3D object pose, where the dense shape and wireframe models are used respectively. In \cite{xiang2015data}, a voxel pattern is employed to detect 3D pose of objects with specific visibility patterns. Similarly to object proposal approaches in 2D detection \cite{ren2015faster}, \cite{chen20153d} generates 3D proposals by utilizing depth information calculated from stereo images, while \cite{chen2016monocular} exploits the ground plane assumption and additional segmentation features to produce 3D candidates; R-CNN is then used for candidates scoring and object recognition. Such high-dimension features used for proposal generating or model fitting are computationally complex for both training and deploying. Instead of directly generating 3D boxes, \cite{mousavian20173d} regresses object orientation and dimensions in separate stages; then the 2D-3D box geometry constraints are used to calculate the 3D object pose, while purely depending on instance 2D box limits its performance in object-truncated cases.

In this work, we study the pros and cons of existing works and propose an integrated perception solution for autonomous driving that makes full use of the instance semantic prior and precise feature spatial-temporal correspondences to achieve robust and continuous state estimation for both the ego-camera and static or dynamic objects in diverse environments.

\section{Overview}
Our semantic tracking system has three main modules, as illustrated in Fig. \ref{fig:system}. The first module performs 2D object detection and viewpoint classification (Sect. \ref{sec:viewpoint}), where the objects poses are roughly inferred based on the constraints between 2D box edges and 3D box vertexes. The second module is feature extraction and matching (Sect. \ref{sec:matching}). It projects all the inferred 3D boxes to the 2D image to get the objects contour and occlusion masks. Guided feature matching is then applied to get robust feature associations for both stereo and temporal images. In the third module (Sect. \ref{sec:estimation}), we integrate all the semantic information, feature measurements into a tightly-coupled optimization approach. A kinematics model is additionally applied to cars to get consistent motion estimation.

\section{Viewpoint Classification and 3D Box Inference}
\label{sec:viewpoint}
Our semantic measurement includes the 2D object box and classified viewpoints. Based on this, the object pose can be roughly inferred instantly in close-form.
\subsection{Viewpoint Classification}
\label{sec:perspective}
\begin{figure}
	\setlength{\belowcaptionskip}{-0.7cm} 
	\centering
	\subfigure[] { \label{fig:perspective} 
		\includegraphics[width=0.41\columnwidth]{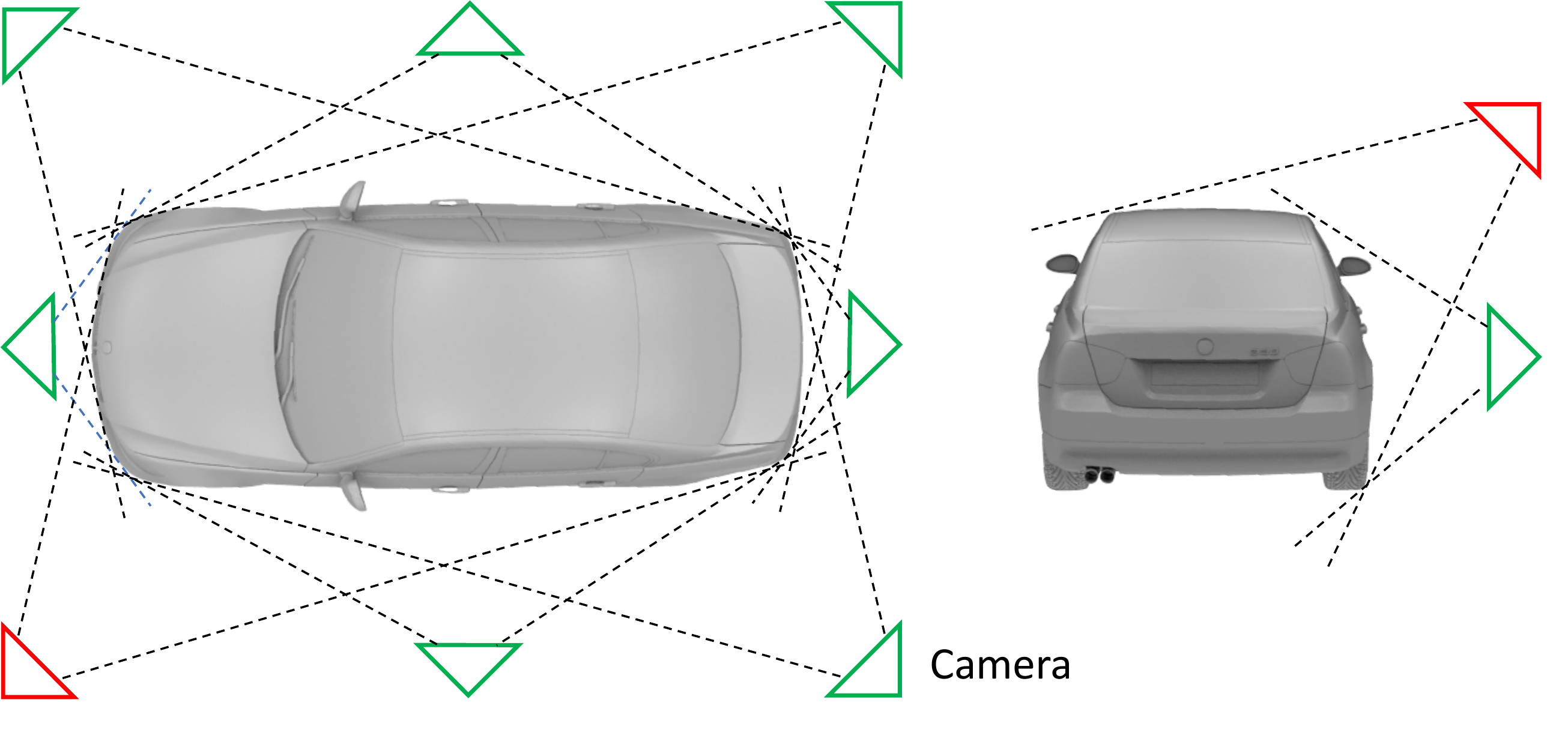}
	}
	\subfigure[] { \label{fig:3dbox} 
		\includegraphics[width=0.41\columnwidth]{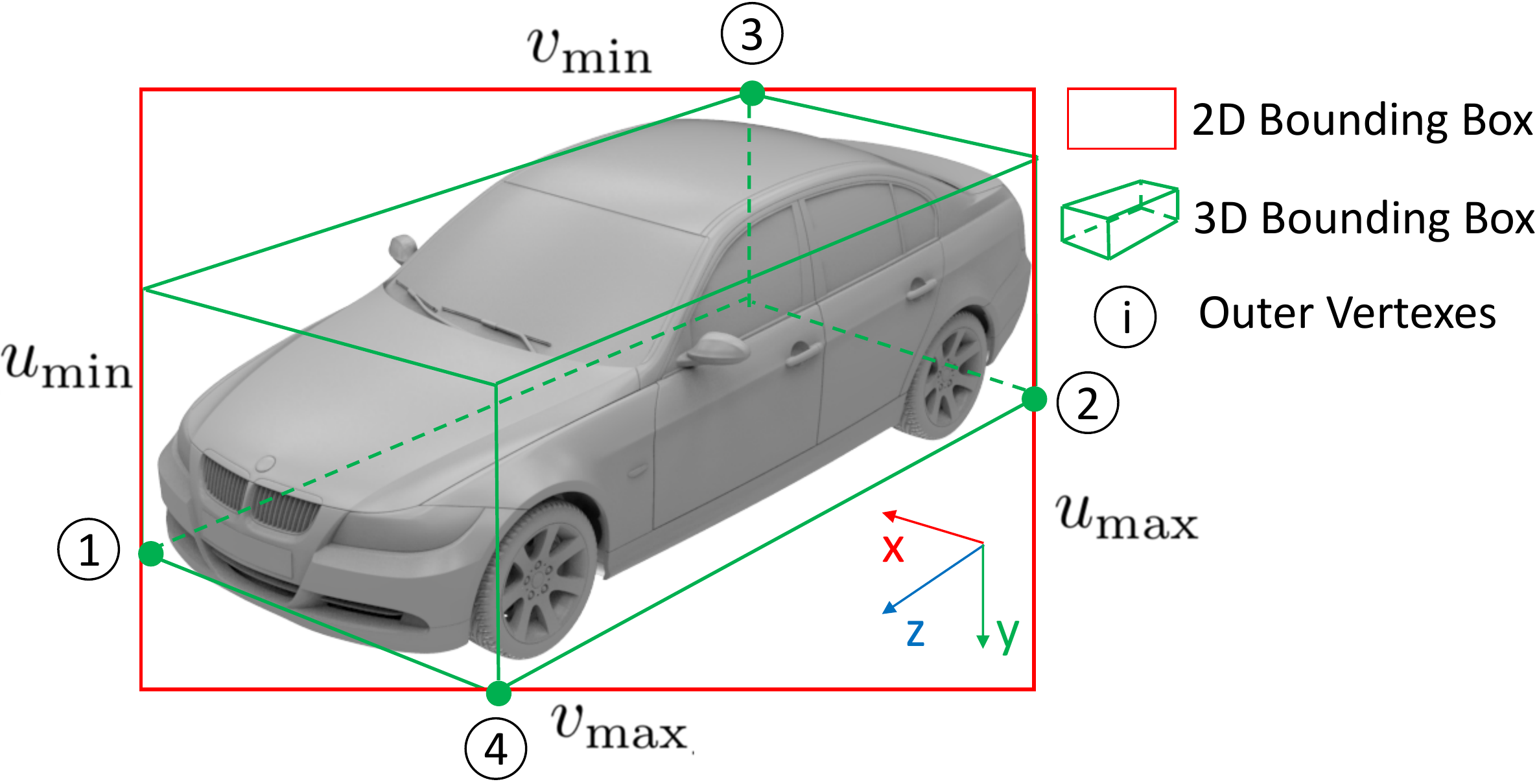}  
	}
	\caption{(a) presents all the horizontal and vertical viewpoints for our classification, their combinations are enough to cover all the observation cases in autonomous scenarios. (b) illustrates the 3D car in a specific viewpoint, where the object frame, four vertexes corresponding to four 2D box edges are denoted respectively.}
\end{figure}

2D Object detection can be implemented by the state-of-the-art object detectors such as Faster R-CNN \cite{ren2015faster}, YOLO \cite{redmon2016you}, etc. We use Faster R-CNN in our system since it performs well on small objects. Only left images are used for object detection due to real-time requirement.
The network architecture is illustrated in Fig. \ref{fig:system} (a). Instead of the pure object classification in the original implementation of \cite{ren2015faster}, we add sub-categories classification in the final FC layers, which denotes object horizontal and vertical discrete viewpoints. 
As Fig \ref{fig:perspective} shown, We divide the continuous object observing angle into eight horizontal and two vertical viewpoints. With total 16 combinations of horizontal and vertical viewpoint classification, 
we can generate associations between edges in the 2D box and vertexes in the 3D box based on the assumption that the reprojection of the 3D bounding box will tightly fit the 2D bounding box.
These associations provide essential condition to build the four edge-vertex constraints for 3D box inference (Sect. \ref{sec:3d-box-pre}) and formulate our semantic measurement model (Sect. \ref{sec:objects-estimation}).

Comparing with direct 3D regression, the well-developed 2D detection and classification networks are more robust over diverse scenarios. The proposed viewpoint classification task is easy to train and have high accuracy, even for small and extreme occluded objects. 
\vspace{-10pt}
\subsection{3D Box Inference Based on Viewpoint}
\label{sec:3d-box-pre}
Given the 2D box described by four edges in normalized image plane $  [u_{\rm min}, v_{\rm min}, \\u_{\rm max}, v_{\rm max}]$ and classified viewpoint, we aim to infer the object pose based on four constriants between 2D box edges and 3D box vertexes, which is inspired by \cite{mousavian20173d}. A 3D bounding box can be represented by its center position $ \mathbf{p} = [p_x, p_y, p_z]^T$ and horizontal orientation $\theta$ respecting to camera frame and the dimensions prior $\mathbf{d} = [d_x, d_y, d_z]^T$. For example, in such a viewpoint presented in Fig. \ref{fig:3dbox} from one of 16 combinations in Fig. \ref{fig:perspective} (denoted as red), four vertexes are projected to the 2D edges, the corresponding constraints can be formulated as:
\begin{equation}
\label{eq:prior}
\left\{  
\begin{array}{lr}  
u_{\rm min} = \pi\left(\mathbf{p} + \mathbf{R}_{\theta}  \mathbf{C}_1  \mathbf{d}\right)_u, \,
u_{\rm max} = \pi\left(\mathbf{p} + \mathbf{R}_{\theta}  \mathbf{C}_2  \mathbf{d}\right)_u, \\
v_{\rm min} = \pi\left(\mathbf{p} + \mathbf{R}_{\theta}  \mathbf{C}_3  \mathbf{d}\right)_v,\,
v_{\rm max} = \pi\left(\mathbf{p} + \mathbf{R}_{\theta}  \mathbf{C}_4  \mathbf{d}\right)_v,  
\end{array}  
\right.  
\end{equation} 
where $\pi$ is a 3D projection warp function which defined as $\pi(\mathbf{p}) = [p_x/p_z, p_y/p_z]^T$, and $(\cdot)_u$ represents the u coordinate in the normalized image plane. We use $\mathbf{R}_{\theta}$ to denote the rotation parameterized by horizontal orientation $\theta$ from the object frame to the camera frame. $\mathbf{C}_{1:4}$ are four diagonal selection matrixes to describe the relations between the object center to the four selected vertexes, which can be determined after we get the classified viewpoint without ambiguous. From the object frame defined in Fig. \ref{fig:3dbox}, it's easy to see that:
\begin{equation}
\label{eq:selection}
\mathbf{C}_{1:4}= \begin{bmatrix}
0.5 &0 &0 \\
0 &0.5 &0\\
0 &0 &0.5
\end{bmatrix},
\begin{bmatrix}
-0.5 &0 &0 \\
0 &0.5 &0\\
0 &0 &-0.5
\end{bmatrix},
\begin{bmatrix}
0.5 &0 &0 \\
0 &-0.5 &0\\
0 &0 &-0.5
\end{bmatrix},
\begin{bmatrix}
-0.5 &0 &0 \\
0 &0.5 &0\\
0 &0 &0.5
\end{bmatrix}.
\end{equation}
With these four equations, the 4 DoF object pose can be solved intuitively given the dimensions prior. This solving process has very trivial time consuming comparing with \cite{mousavian20173d} which exhaustive tests all the valid edge-vertex configurations.

We convert the complex 3D object detection problem into 2D detection, viewpoint classification, and straightforward closed-form calculation. Admittedly, the solved pose is an approximated estimation which is conditioned on the instance "tightness" of the 2D bounding box and the object dimension prior. Also for some top view cases, the reprojection of the 3D box does not strictly fit the 2D box, which can be observed from the top edge in Fig. \ref{fig:3dbox}. However, for the almost horizontal or slight looking-down viewpoints in autonomous driving scenarios, this assumption can be held reasonably. Note that our instance pose inference is only for generating object projection contour and occlusion mask for the feature extraction (Sect. \ref{sec:matching}) and serves as an initial value for the follow-up maximum-a-posteriori (MAP) estimation, where the 3D object trajectory will be further optimized by sliding window based feature correlation and object point cloud alignment.

\section{Feature Extraction and Matching}
\label{sec:matching}
We project the inferred 3D object boxes (Sect. \ref{sec:3d-box-pre}) to the stereo images to generate a valid 2D contour. As Fig. \ref{fig:system} (b) illustrates, we use different colors mask to represent visible part of each object (gray for the background). For occlusion objects, we mask the occluded part as invisible according to objects 2D overlap and 3D depth relations. For truncated objects which have less than four valid edges measurements thus cannot be inferred by the method in Sec. \ref{sec:3d-box-pre}, we directly project the 2D box detected in the left image to the right image. We extract ORB features \cite{rublee2011orb} for both the left and right image in the visible area for each object and the background.

Stereo matching is performed by epipolar line searching. The depth range of object features are known from the inferred object pose, so we limit the search area to a small range to achieve robust feature matching. For temporal matching, we first associate objects for successive frames by 2D box similarity score voting. The similarity score is weighted by the center distance and shape similarity of the 2D boxes between successive images after compensating the camera rotation. The object is treated as lost if its maximum similarity score with all the objects in the previous frame is less than a threshold. We note that there are more sophisticated association schemes such as probabilistic data association \cite{bowman2017probabilistic}, but it is more suitable for avoiding the hard decision when re-visiting the static object scene than for the highly dynamic and no-repetitive scene for autonomous driving. We subsequently match ORB features for the associated objects and background with the previous frame. Outliers are rejected by RANSAC with local fundamental matrix test for each object and background independently.

\section{Ego-motion and Object Tracking}
\label{sec:estimation}
Beginning with the notation definition, we use $\mathbf{s}^t_{\rm k} = \{{\mathbf{b}}^t_{kl}, \mathbf{b}^t_{kr}, l_{k}, {\mathbf{C}}^t_{k1:4}\}$ to denote the semantic measurement of the $k^{th}$ object at time $t$, where ${\mathbf{b}}^t_{kl}, \mathbf{b}^t_{kr}$ are the observations of the left-top and the right-bottom coordinates of the 2D bounding box respectively, $l_{k}$ is the object class label and ${\mathbf{C}}^t_{k1:4}$ are four selection matrixes defined in Sec. \ref{sec:3d-box-pre}. For measurements of sparse feature which is anchored to one object or the background, we use $^n{\mathbf{z}}^t_{k} = \{{^n}\mathbf{z}^t_{kl}, {^n}\mathbf{z}^t_{kr}\}$ to denote the stereo observations of the $n^{th}$ feature on the $k^{th}$ object at time $t$ ($k = 0$ for the static background), where ${^n}\mathbf{z}^t_{kl}, {^n}\mathbf{z}^t_{kr}$ are  feature coordinates in the normalized left and right \begin{wrapfigure}{r}{0.4\textwidth}
	\setlength{\belowcaptionskip}{-0.7cm}
	\includegraphics[width=0.9\linewidth]{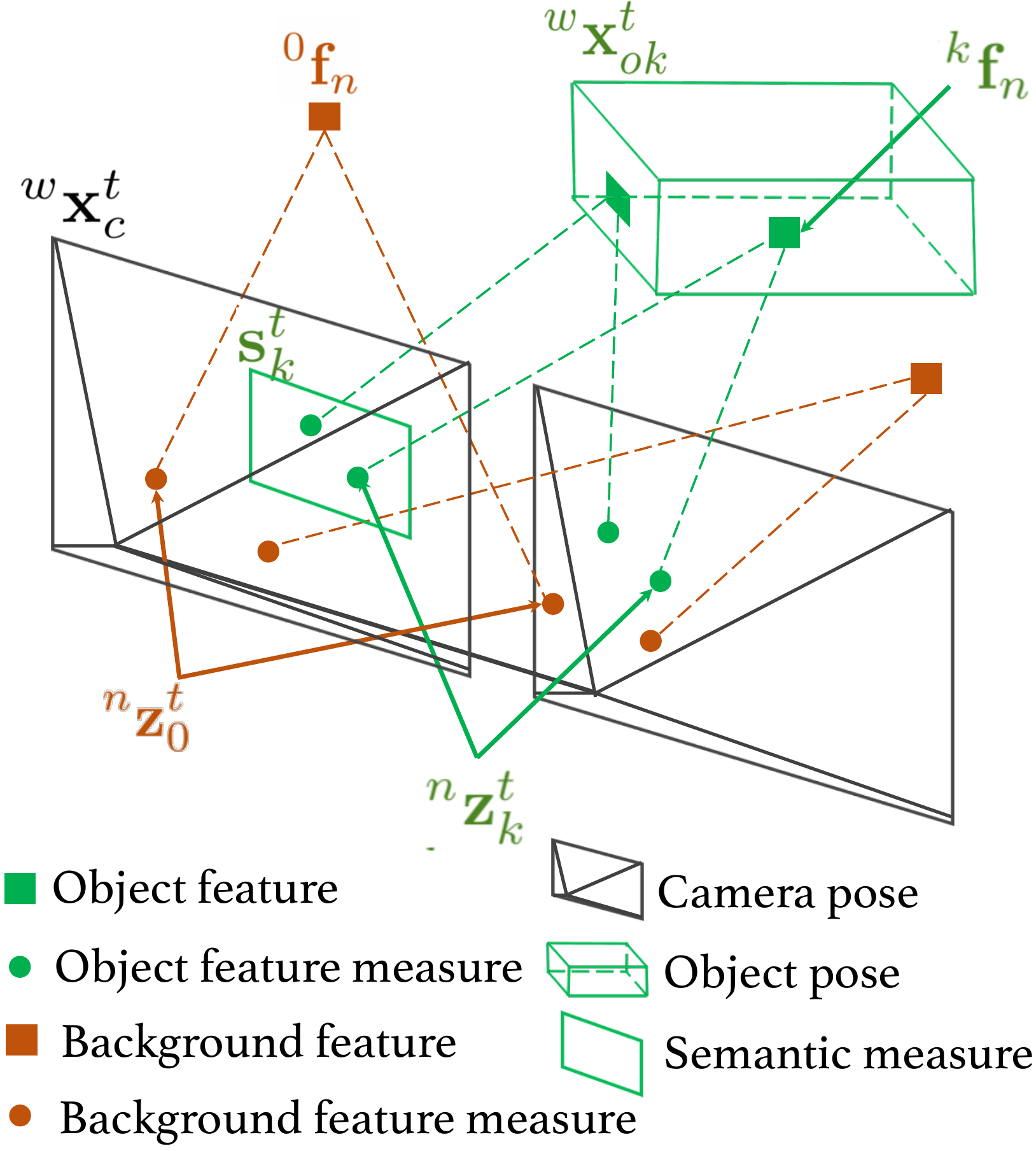} 
	\caption{Notation visualization.}
	\label{fig:notation}
\end{wrapfigure}image plane respectively. The states of the ego-camera and the $k^{th}$ object are represented as $^w\mathbf{x}^t_c = \{{^w}\mathbf{p}^t_c, {^w}\mathbf{R}^t_c\}$, ${^w}\mathbf{x}^t_{ok} = \{{^w}\mathbf{p}^t_{ok}, \mathbf{d}_k, {^w}{{\theta}}^t_{ok}, v^t_{ok}, \delta^t_{ok} \}$ respectively, 
where we use $^w(\cdot)$, $(\cdot)_c$ and $(\cdot)_o$ to denote the world, camera and object frame. $^w\mathbf{p}$ represents the position in the world frame. 
For objects orientation, we only model the horizontal rotation $^w{\theta}^t_{ok}$  instead of $\mathbb{SO}(3)$ rotation $^w{\mathbf{R}}^t_{c}$ for the ego-camera. $\mathbf{d}_{k}$ is the time-invariant dimensions of the $k^{th}$ object, and $v^t_{ok}, \delta^t_{ok}$ are the speed and steering angle, which are only estimated for cars. For conciseness, we visualize the measurements and states in Fig. \ref{fig:notation} at the time $t$. 

Considering a general autonomous driving scene, we aim to continuously estimate the ego-motion of the onboard camera from time $0$ to $T$: $^w\mathcal{X}_c = \{{^w\mathbf{x}^t_c}\}_{t=0:T}$, and track the $K_t$ number of 3D objects: $^w\mathcal{X}_o = \left\{ {^w\mathbf{x}_{ok}} \right \}_{k=1:K_t}, {^w\mathbf{x}_{ok}} = \{{^w\mathbf{x}^t_{ok}}\}_{t=0:T},$
and recover the 3D position of the dynamic sparse features: $\mathcal{F} = \{^k\mathbf{f}\}_{k=0:K_t}, {^k}\mathbf{f} = \{^k{\mathbf{f}}_n\}_{n=0:N_k}$, (note that here we use $^k(\cdot)$ to denote the $k^{th}$ object frame, in which the features are relatively static, $k=0$ for background world, in which the features are globally static), given the semantic measurements: $\mathcal{S} = \{\mathbf{s}_k\}_{k=1:K_t},\, \mathbf{s}_k = \{\mathbf{s}^t_k\}_{t=0:T}$ and sparse feature observations anchored to the $k^{th}$ object: $\mathcal{Z} = \{\mathbf{z}_k\}_{k=0:K_t}, \,\mathbf{z}_k = \{{^n}\mathbf{z}_{k}\}_{n=0:N_k}.
{^n}\mathbf{z}_{k} = \{{^n}{\mathbf{z}}^t_{k}\}_{t= 0:T}$.
We formulate our semantic objects and camera ego-motion tracking from the probabilistic model to a nonlinear optimization problem.

\subsection{Ego-motion Tracking}
Given the static background feature observation, the ego-motion can be solved via maximum likelihood estimation (MLE):
\begin{align}
\label{eq:ego-motion2}
^w\mathcal{X}_c, {^0}\mathbf{f} &= \argmaxA_{{^w}\mathcal{X}_c, {^0}\mathbf{f}} \prod^{N_0}_{n=0}\prod^T_{t=0} p({^n}\mathbf{z}^t_0|^w\mathbf{x}^t_c, {^0}{\mathbf{f}_n}, {^w}\mathbf{x}^{0}_c) \; \\ 
&= \argmaxA_{^w\mathcal{X}_c, {^0}\mathbf{f}} \sum^{N_0}_{n=0}\sum^T_{t=0} \log p({^n}\mathbf{z}^t_0|^w\mathbf{x}^t_c, {^0}{\mathbf{f}_n}, {^w}\mathbf{x}^{0}_c) \; \\
&= \argminA_{^w\mathcal{X}_c, {^0}\mathbf{f}} \sum^{N_0}_{n=0}\sum^T_{t=0} \left \| r_{\mathcal{Z}}(^n{\mathbf{z}}^{t}_{0},\, ^w\mathbf{x}^t_c , {^0}{\mathbf{f}_n}) \right \|_{^0\mathbf{\sum}^{t}_{n}}^2.
\end{align}
This is the typical SLAM or SfM approach. The camera pose and background point cloud are estimated conditionally on the first state.
 As Eq. \ref{eq:ego-motion2} shows, the log probability of measurement residual is proportional to its Mahalanobis norm $\left \| \mathbf{r} \right\|_{\scriptstyle{\mathbf{\sum}}}^2 = \mathbf{r}^T {\scriptstyle{\sum}}^{-1}{\mathbf{r}}$. Then the MLE is converted to a nonlinear least square problem, this process is also known as Bundle Adjustment(BA).
\subsection{Semantic Object Tracking}
\label{sec:objects-estimation}
After we solve the camera pose, the object state at each time $t$ can be solved based on the dimension prior and the instance semantic measurements (Sect \ref{sec:3d-box-pre}). We assume the object is a rigid body, which means the feature anchored to it is fixed respecting to the object frame. Therefore, the temporal states of the object are correlated if we have continuous object feature observations. States of different objects are conditionally independent given the camera pose, so we can track all the objects in parallel and independently. For the $k^{th}$ object, we have the dimension prior distribution $p(\mathbf{d}_{k})$ for each class label. We assume the detection results and feature measurements for each object at each time are independent and Gaussian distributed. According to Bayes' rule, we have the following maximum-a-posteriori (MAP) estimation:
\begin{align}
\label{eq:object-motion3}
^w\mathbf{x}_{ok}, {^k}\mathbf{f} = &\argmaxA_{^w\mathbf{x}_{ok}, {^k}\mathbf{f}} p(^w\mathbf{x}_{ok}, {^k}\mathbf{f} \,|\, {^w}\mathbf{x}_{c}, \mathbf{z}_k, \mathbf{s}_k) \\
= & \argmaxA_{^w\mathbf{x}_{ok}, {^k}\mathbf{f}} p(\mathbf{z}_k, \mathbf{s}_k | {^w}\mathbf{x}_{c}, {^w}\mathbf{x}_{ok}, {^k}\mathbf{f}) p(\mathbf{d}_{k}) \; \\
= & \argmaxA_{^w\mathbf{x}_{ok}, {^k}\mathbf{f}} p(\mathbf{z}_k | {^w}\mathbf{x}_{c}, {^w}\mathbf{x}_{ok}, {^k}\mathbf{f}) p(\mathbf{s}_k | {^w}\mathbf{x}_{c}, {^w}\mathbf{x}_{ok}) p(\mathbf{d}_{k}) \\
= \argmaxA_{^w\mathbf{x}_{ok}, {^k}\mathbf{f}} \prod^T_{t=0} \prod^{N_k}_{n=0} &p({^n}{\mathbf{z}}^t_k | {^w}\mathbf{x}^t_{c}, {^w}\mathbf{x}^t_{ok}, {^k}{\mathbf{f}}_n) p(\mathbf{s}^t_k | {^w}\mathbf{x}^t_{c}, {^w}\mathbf{x}^t_{ok}) p({^w}\mathbf{x}^{t-1}_{ok} | {^w}\mathbf{x}^{t}_{ok}) p(\mathbf{d}_{k}).
\end{align}
Similar to Eq. \ref{eq:ego-motion2}, we convert the MAP to a nonlinear optimization problem:

\begin{align}
\label{eq:object-motion2}
^w\mathbf{x}_{ok}, {^k}\mathbf{f} =& \argminA_{^w\mathbf{x}_{ok}, {^k}\mathbf{f}} \left\{ \sum^T_{t=0} \sum^{N_k}_{n=0} \left \| r_{\mathcal{Z}}(^n{\mathbf{z}}^{t}_{k},\, {^w}\mathbf{x}^t_c ,{^w}\mathbf{x}^t_{ok}, {^k}{\mathbf{f}_n}) \right \|_{^k\mathbf{\sum}^{t}_{n}}^2 \right. + \left \| r_{\mathcal{P}}(d^l_k, \mathbf{d}_{k}) \right \|_{\mathbf{\sum}^{l}}^2 \nonumber \\ 
& \,\,+ \sum^{T}_{t=1} \left \| r_{\mathcal{M}}({^w}\mathbf{x}^{t}_{ok}, {^w}\mathbf{x}^{t-1}_{ok}) \right \|_{\mathbf{\sum}^{t}_{k}}^2 \left. +\sum^T_{t=0} \left \| r_{\mathcal{S}}({\mathbf{s}}^{t}_{k},\, {^w}\mathbf{x}^t_c ,{^w}\mathbf{x}^t_{ok}) \right \|_{\mathbf{\sum}^{t}_{k}}^2 \right\},
\end{align}
where we use $r_\mathcal{Z}$, $r_\mathcal{P}$, $r_\mathcal{M}$, and $r_\mathcal{S}$ to denote the residual of the feature reprojection, dimension prior, object motion model, and semantic bounding box reprojection respectively. $\scriptstyle{\mathbf{\sum}}$ is the corresponding covariance matrix for each measurement. We formulate our 3D object tracking problem into a dynamic object BA approach which makes fully exploit object dimension and motion prior and enforces temporal consistency. Maximum a posteriori estimation can be achieved by minimizing the sum of the Mahalanobis norm of all the residuals.

\subsubsection{Sparse Feature Observation} We extend the projective geometry between static features and camera pose to dynamic features and object pose. Based on anchored relative static features respecting to the object frame, the object poses which share feature observations can be connected by a factor graph. For each feature observation, the residual can be represented by the reprojection error of predicted feature position and the actual feature observations on the left and right image:
\begin{align}
\label{eq:feature-factor}
&r_{\mathcal{Z}}(^n\mathbf{z}^{t}_{k},\, {^w}\mathbf{x}^t_c ,{^w}\mathbf{x}^t_{ok}, {^k}{\mathbf{f}_n}) \\ =& 
\begin{bmatrix}
\pi\left(h^{-1}({^w}\mathbf{x}^t_c,\, h({^w}\mathbf{x}^t_{ok}, {^k}{\mathbf{f}_n}))\right) - {^n}{\mathbf{z}}^{t}_{kl} \\
\pi\left(h({^{r}}\mathbf{x}_l, h^{-1}({^w}\mathbf{x}^t_c,\, h({^w}\mathbf{x}^t_{ok}, {^k}{\mathbf{f}_n})))\right) - {^n}{\mathbf{z}}^{t}_{kr} 
\end{bmatrix},  
\end{align}
where we use $h(\mathbf{x}, p)$ to denote applying a 3D rigid body transform $\mathbf{x}$ to a point $p$. For example, $h\left({^w}\mathbf{x}^t_{ok}, {^k}{\mathbf{f}_n}\right)$ transforms the $n^{th}$ feature point ${^k}{\mathbf{f}_n}$ from the object frame to the world frame,  $h^{-1}(\mathbf{x}, p)$ is the corresponding inverse transform. ${^r}\mathbf{x}_l$ denotes the extrinsic transform of the stereo camera, which is calibrated offline. 

\subsubsection{Semantic 3D Object Measurement} Benefiting from the viewpoint classification, we can know the relations between edges of the 2D bounding box and vertexes of the 3D bounding box. Assume the 2D bounding box is tightly fitted to the object boundary, then each edge is intersected with a reprojected 3D vertex. These relations can be determined as four selection matrixes for each 2D edge. The semantic residual can be represented by the reprojection error of the predicted 3D box vertexes with the detected 2D box edges: 
\begin{align}
\label{eq:semantic-factor}
&r_{\mathcal{S}}({\mathbf{s}}^{t}_{k},\, {^w}\mathbf{x}^t_c ,{^w}\mathbf{x}^t_{ok}, {\mathbf{d}_{k}})
= \begin{bmatrix}
\pi\left(h_{\mathbf{C}_1}\right)_u - ({\mathbf{b}}^{t}_{kl})_u \\
\pi\left(h_{\mathbf{C}_2}\right)_u - ({\mathbf{b}}^{t}_{kr})_u \\
\pi\left(h_{\mathbf{C}_3}\right)_v - ({\mathbf{b}}^{t}_{kl})_v \\
\pi\left(h_{\mathbf{C}_4}\right)_v - ({\mathbf{b}}^{t}_{kr})_v
\end{bmatrix},   \\
& h_{\mathbf{C}_i} = h^{-1}({^w}\mathbf{x}^t_c,\, h({^w}\mathbf{x}^t_{ok}, \mathbf{C}_i{\mathbf{d}^l_k})), 
\end{align}
where we use $h_{\mathbf{C}_i}$ to project a vertex specified by the corresponding selection matrix $\mathbf{C}_i$ of the 3D bounding box to the camera plane. This factor builds the connection between the object pose and its dimensions instantly. Note that we only perform 2D detection on the left image due to the real-time requirement.

\subsubsection{Vehicle Motion Model}
To achieve consistent estimation of motion and orientation for the vehicle class, we employ the kinematics model introduced in \cite{gu2017improved}. The vehicle state at time $t$ can be predicted with the state at ${t-1}$:  
\begin{align}
{^w}\hat{\mathbf{x}}^{t}_{ok} = \begin{bmatrix}
{^w}\mathbf{p}^t_{ok}  \\
{^w}{{\theta}}^t_{ok} \\
\delta^{t}_{ok}\\
v^{t}_{ok}
\end{bmatrix} & = \begin{bmatrix}
\mathbf{I}_{3 \rm x 3}, & \mathbf{0}, & \mathbf{0}, & \boldsymbol{\Lambda} \\
\mathbf{0}, & 1, &0, &\frac{\rm{tan}(\delta) \Delta t}{L} \\
\mathbf{0}, &0, &1, &0 \\
\mathbf{0}, &0, &0, &1
\end{bmatrix} \begin{bmatrix}
{^w}\mathbf{p}^{t-1}_{ok}  \\
{^w}{{\theta}}^{t-1}_{ok} \\
\delta^{t-1}_{ok} \\    
v^{t-1}_{ok}
\end{bmatrix}, \boldsymbol{\Lambda} = \begin{bmatrix}
\rm{cos}(\theta)\Delta t \\
\rm{sin}(\theta)\Delta t\\
0
\end{bmatrix}, \\
&r_{\mathcal{M}}({^w}{\mathbf{x}}^{t}_{ok},{^w}\mathbf{x}^{t-1}_{ok}) = {^w}{\mathbf{x}}^{t}_{ok} - {^w}\hat{\mathbf{x}}^{t}_{ok},
\end{align}
where $L$ is the length of the wheelbase, which can be parameterized by the dimensions. The orientation of the car is always parallel to the moving direction. We refer readers to \cite{gu2017improved} for more derivations. Thanks to this kinematics model, we can track the vehicle velocity and orientation continuously, which provides rich information for behavior and path planning for autonomous driving. For other class such as pedestrians, we directly use a simple constant-velocity model to enhance smoothness.
\subsubsection{Point Cloud Alignment}
\begin{wrapfigure}{r}{0.4\textwidth}
	\vspace{-0.8cm}
	\setlength{\belowcaptionskip}{-0.5cm}
	\includegraphics[width=0.9\linewidth]{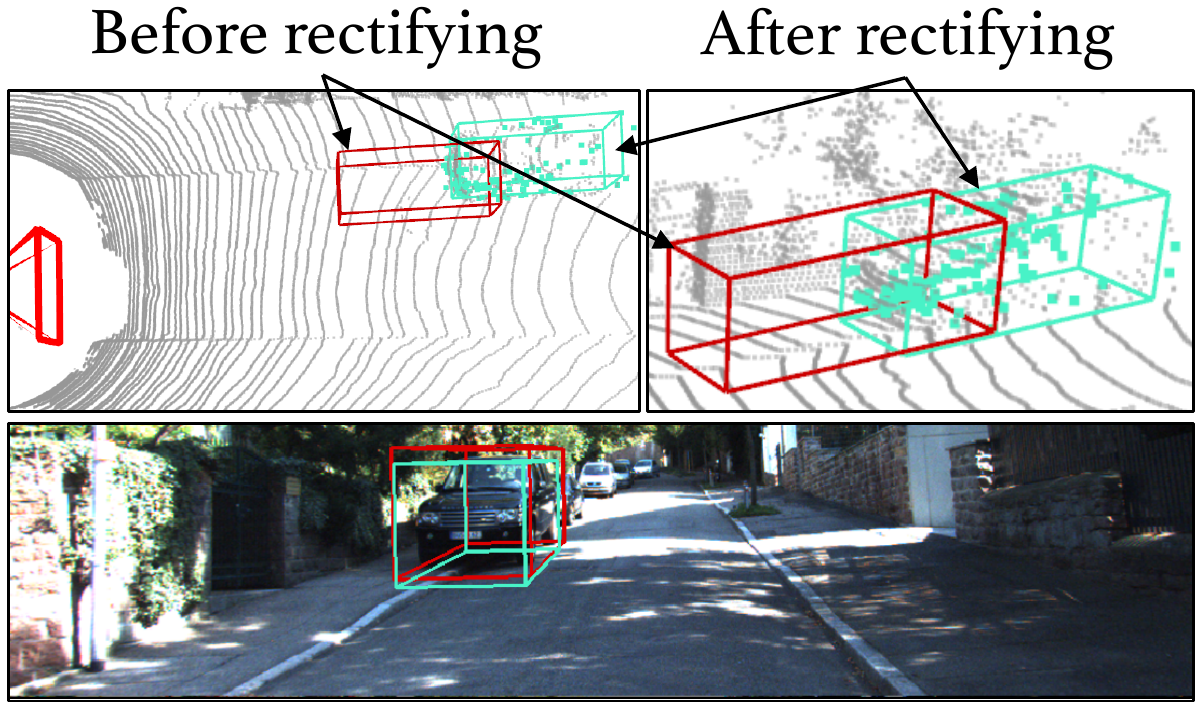}
	\caption{Point cloud alignment.}
	\label{fig:rectify}
\end{wrapfigure}After minimizing all the residuals, we obtain the MAP estimation of the object pose based on the dimension prior. However, the pose might be biased estimated due to object size difference (See Fig. \ref{fig:rectify}). We therefore align the 3D box to the recovered point cloud, which is unbiased because of accurate stereo extrinsic calibration. We minimize the distance of all 3D points with their anchored 3D box surfaces:
\begin{align}
^w{\mathbf{x}}^{t}_{ok} = \argminA_{^w\mathbf{x}_{ok}} \sum^{N_k}_{n=0} d({^w\mathbf{x}}^{t}_{ok}, ^k\mathbf{f}_n),
\end{align}
where $d({^w\mathbf{x}}^{t}_{ok}, ^k\mathbf{f}_n)$ denotes the distance of the $k^{th}$ feature with its corresponding observed surface. After all the above information is tightly fused together, we get consistent and accurate pose estimation for both the static and dynamic objects.
\begin{figure}
	\centering
	\setlength{\belowcaptionskip}{-1cm} 
	\subfigure[] { \label{fig:long} 
		\includegraphics[width=1.0\columnwidth]{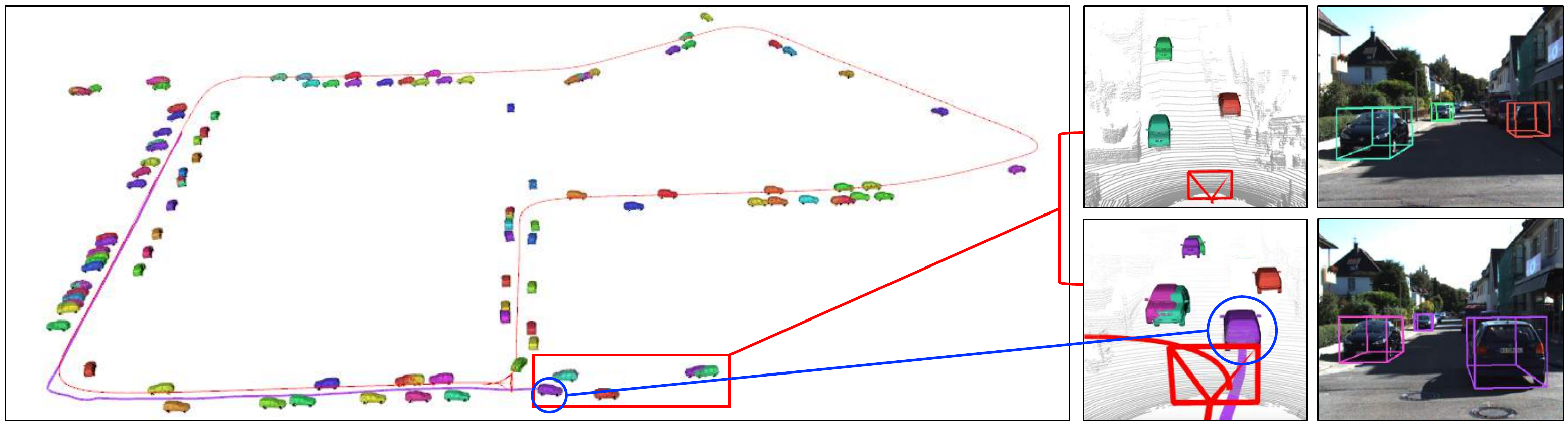}
	} 
	\subfigure[] { \label{fig:track} 
		\includegraphics[width=1.0\columnwidth]{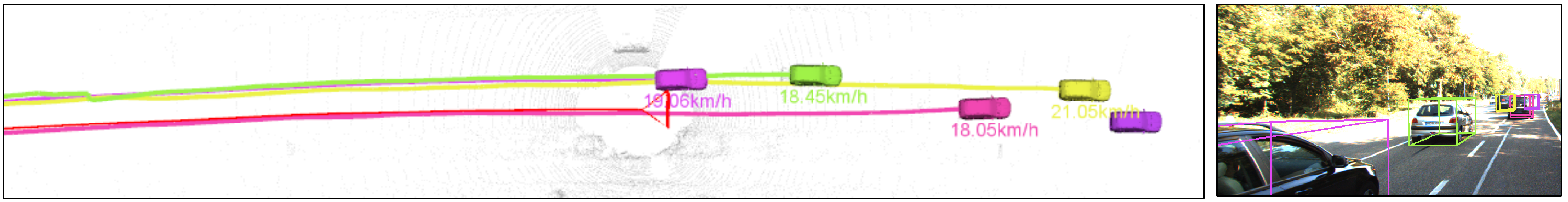}
	}
	\caption{Continuous tracking results over long trajectories. (a) shows a roughly 700 m close-loop trajectory including both static and dynamic cars. The right top and right bottom are enlarged start and end views respectively. The car in the blue circle is tracked over 200 meters, the trajectory of which can be found in the left top view. (b) shows a scenario which mainly contains dynamic and truncated cars. The estimated trajectory, velocity and reprojected 2D image are presented in left and right respectively. Note that the LiDAR point cloud is only for reference in all the top views.
		\label{fig:long_tra}}
\end{figure}
\section{Experimental Results}
We evaluate the performance of the proposed system on both KITTI \cite{Geiger2012CVPR,Geiger2013IJRR} and Cityscapes \cite{cordts2015cityscapes} dataset over diverse scenarios. The mature 2D detection and classification module has good generalization ability to run on unseen scenes, and the follow-up nonlinear optimization is data-independent. Our system is therefore able to perform consistent results on different datasets. The quantitative evaluation shows our semantic 3D object and ego-motion tracking system has better performance than the isolated state-of-the-art solutions.

\subsection{Qualitative Results Over Diverse Scenarios}
\begin{figure}
	\setlength{\belowcaptionskip}{-0.5cm}
	\centering
	\includegraphics[width=1.0\columnwidth]{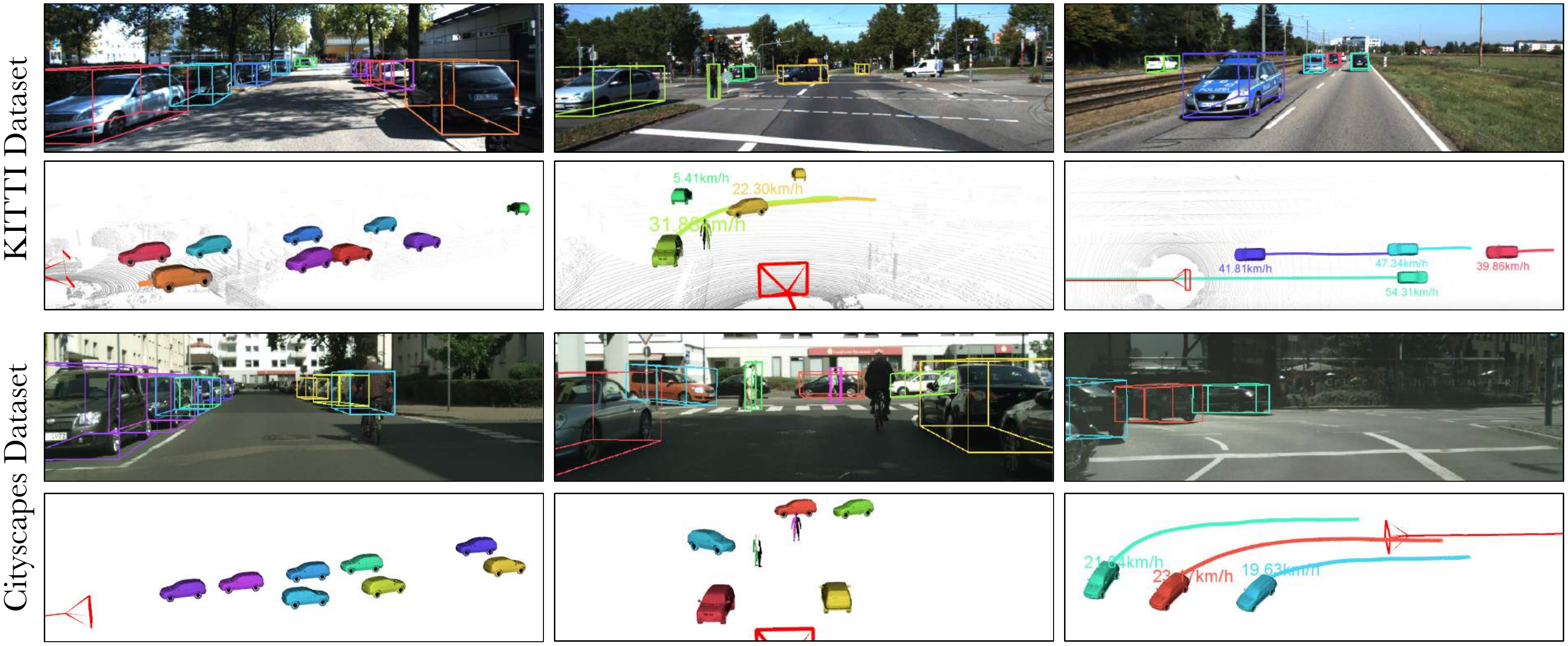}
	\caption{Qualitative examples over diverse scenarios. From left to right: Concentrated cars. Crossroads which include both cars and pedestrians (note that we do not solve orientation for pedestrians), Dynamic cars. The top two rows are the results on the KITTI dataset, and the bottom two rows show the results on the Cityscapes dataset.
		\label{fig:quanlitative}}
\end{figure}

Firstly, we test the system on long challenging trajectories in KITTI dataset which contains 1240 $\times$ 376 stereo color and grayscale images captured at 10 Hz. We perform 2D detection on left color images and extract 500 (for the background) and 100 (for the object) ORB features \cite{rublee2011orb} on both left and right grayscale images. Fig. \ref{fig:long} shows a 700 m close-loop trajectory which includes both static and dynamic cars. We use red cone and line to represent the camera pose and trajectory, and various color CAD models and lines to represent different cars and their trajectories, all the observed cars are visualized in the top view. 
Currently, our system performs object tracking in a memoryless manner, so the re-observed object will be treated as a new one, which can also be found in the enlarged start and end views in Fig. \ref{fig:long}. In Fig. \ref{fig:track}, the black car is continuously truncated over a long time, which is an unobservable case for instance 3D box inference (Sect. {\ref{sec:3d-box-pre}}). However, we can still track its pose accurately due to temporal feature constraints and dynamic point cloud alignment.

We also demonstrate the system performance on different datasets over more scenarios which include concentrated  cars, crossroads, and dynamic roads. All the reprojected images and the corresponding top views are shown in Fig. \ref{fig:quanlitative}.

\subsection{Quantitative Evaluation}
Since there are no available integrated academic solutions for both ego-motion and dynamic objects tracking currently, we conduct quantitative evaluations for the camera and objects poses by comparing with the isolated state-of-the-art works: ORB-SLAM2 \cite{murORB2} and 3DOP \cite{chen20153d}.
\begin{figure}
	\setlength{\belowcaptionskip}{-1cm} 
	\centering
	\includegraphics[width=1.0\columnwidth]{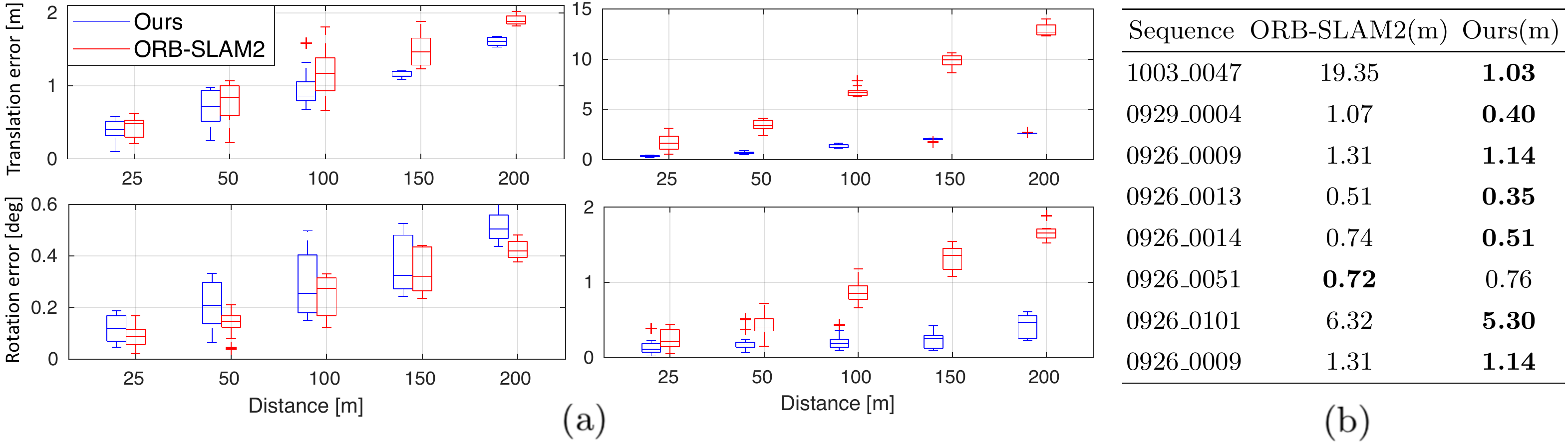}
	\caption{(a) RPEs comparison. Left and right are the results of 0929\_0004 and 1003\_0047 sequences from the KITTI raw dataset respectively. (b) RMSEs of ATE comparisons on ten long KITTI raw sequences.
		\label{fig:vo}}
\end{figure}
\subsubsection{Camera Pose Evaluation}
Benefiting from the semantic prior, our system can perform robust camera estimation in dynamic environments. We evaluate the accuracy of camera odometry by comparing the relative pose error (RPE) \cite{Geiger2012CVPR} and RMSE of ATE (Absolute Trajectory Error) \cite{sturm2012benchmark} with the ORB-SLAM2 \cite{murORB2} with stereo settings. Two sequences in KITTI raw dataset: {0929\_0004} and {1003\_0047} which include dynamic objects are used for RPEs comparison. The relative translation and rotation errors are presented in Fig. \ref{fig:vo} (a). Ten long sequences of KITTI raw dataset are additionally used to evaluate RMSEs of ATE, as detailed in Fig. \ref{fig:vo} (b). It can be seen that our estimation shows almost same accuracy with \cite{murORB2} in less dynamic scenarios due to the similar Bundle Adjustment approaches (0926\_0051, etc.). However, our system still works well in high dynamic environments while ORB-SLAM2 shows non-trivial errors due to introducing many outliers (1003\_0047, 0929\_0004, etc.). This experiment shows that the semantic-aided object-aware property is essential for camera pose estimation, especially for dynamic autonomous driving scenarios.

\subsubsection{Object Localization Evaluation}
We evaluate the car localization performance on KITTI tracking dataset since it provides sequential stereo images 
with labeled objects 3D boxes. According to occlusion level and 2D box height, we divide all the detected objects into three regimes: easy, moderate and hard then 
evaluate them separately. To evaluate the localization accuracy of the proposed 
estimator, we collect objects average position error statistics. By setting series of 
\begin{wrapfigure}{r}{0.5\textwidth}
	\setlength{\belowcaptionskip}{-0.4cm} 
	\includegraphics[width=1.0\linewidth]{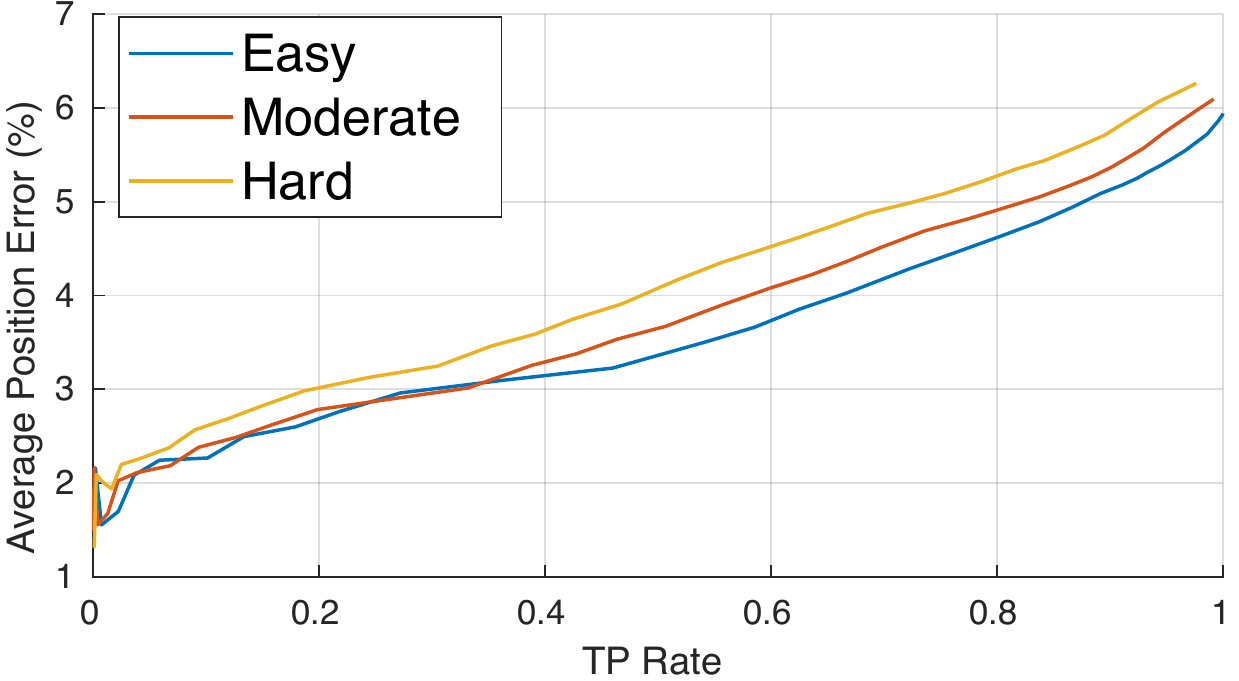} 
	\caption{Average position error vs TP rate results. We set 40 discrete IoU thresholds from 0 to 1, then count the TP rate and the average position error for the true positives for each IoU threshold.}
	\label{fig:error}
\end{wrapfigure}Intersection-over-Unions (IoU) thresholds from 0 to 1, we calculate the true positive (TP) rate and the average error between estimated position of TPs and ground truth at each instance frame for each threshold. The average position error (in \%) vs TP rate curves are shown in Fig. \ref{fig:error}, where we use blue, red, yellow lines to represent statistics for easy, moderate and hard objects. It can be seen that the average error for half tuth positive objects is below 5\%. For all the estimated results, the average position errors are 5.9\%, 6.1\% and 6.3\% for easy, moderate and hard objects respectively.

To compare with baselines, we evaluate the Average Precision (AP) for bird's eye view boxes and 3D boxes by comparing with 3DOP \cite{chen20153d}, the state-of-the-art stereo based 3D object detection method. We set IoU thresholds to 0.25 and 0.5 for both bird's eye view and 3D boxes. Note that we use the oriented box overlap, so the object orientation is also implied evaluated in these two metrics. We use S, M, F, P to represent semantic measurement, motion model, feature observation, and point cloud alignment respectively. As listed in Table.1, the semantic measurement serves as the basis of the 3D object estimation. Adding feature observation increases the AP for easy (near) objects obviously due to large feature extraction area (same case for adding point clout alignment), while adding motion model helps the hard (far) objects since it "smooths" the non-trivial 2D detection noise for small objects. After integrating all these cues together, we obtain accurate 3D box estimation for both near and far objects. It can be seen that our integrated method shows more accurate results for all the AP in bird's eye view and 3D box with 0.25 IoU threshold. Due to the unregressed object size, our performance slightly worse than 3DOP in 3D box comparison of 0.5 IoU. However, we stress our method can efficiently track both static and dynamic 3D objects with temporal smoothness and motion consistency, which is essential for continuous perception and planning in autonomous driving.
\begin{table}
	\begin{center}
		\caption{Average precision (in \%) of bird's eye view and 3D boxes comparison.}
		\label{table:3dbox}
		\renewcommand{\arraystretch}{1.3}
		\resizebox{0.95\textwidth}{!}{%
			\begin{tabular}{c|ccc|ccc|ccc|ccc|c}
				\hline
				\multirow{2}{*}{Method} &
				\multicolumn{3}{c}{$\rm AP_{bv}$(IoU=0.25)} & \multicolumn{3}{c|}{$\rm AP_{bv}$(IoU=0.5)} & \multicolumn{3}{c}{$\rm AP_{3d}$(IoU=0.25)} & \multicolumn{3}{c|}{$\rm AP_{3d}$(IoU=0.5)} & \multirow{2}{*}{Time}\\
				\cline{2-13}
				& \,Easy\, & Mode & \,Hard\, &  \,Easy\, & Mode & \,Hard\, & \,Easy\, & Mode & \,Hard\, &  \,Easy\, & Mode & \,Hard \, & (ms)\\
				\hline
				S  & 63.12 & 56.37 & 53.18 & 33.12 & 28.91 & 27.77 & 58.78 & 52.42 & 48.82 & 25.68 & 21.70 & 21.02 & 120\, \\
				S+M & 66.27 & 63.81 & 58.84 & 41.08 & 38.90 & 34.84 & 62.97 & 60.70 & 55.28 & 34.18 & 30.98 & 27.32 & 121\,\\
				S+F & 76.23 & 70.18 & 66.18 & 48.82 & 43.07 & 39.80 & 73.35 & 66.86 & 62.66 & 38.93 & 33.43 & 30.46 & 170\, \\
				S+F+M & {77.87} & 74.48 & 70.85 & 46.96 & 44.39 & 42.23 & 73.32 & 71.06 & 67.30 & 40.50 & 36.28 & 34.59 & 171\, \\
				S+F+M+P & \textbf{88.07} & \textbf{77.83} & \textbf{72.73} & \textbf{58.52} & \textbf{46.17} & \textbf{43.97} & \textbf{86.57} & \textbf{74.13} & \textbf{68.96} & {48.51} & 37.13 & 34.54 & 173\,\\
				\cline{1-14}
				3DOP & 81.34 & 70.70 & 66.32 & 54.83 & 43.36 & 37.15 & 80.62 & 70.01 & 65.76 & \textbf{53.73} & \textbf{42.27} & \textbf{35.87} & 1200\,\\
				\hline
		\end{tabular}}
	\end{center}
\end{table}
\section{Conclusions and Future work}
In this paper, we propose a 3D objects and ego-motion tracking system for autonomous driving. We integrate the instance semantic prior, sparse feature measurement and kinematics motion model into a tightly-coupled optimization framework. Our system can robustly estimate the camera pose without being affected by the dynamic objects and track the states and recover dynamic sparse features for each observed object continuously. Demonstrations over diverse scenarios and different datasets illustrate the practicability of the proposed system. Quantitative comparisons with state-of-the-art approaches show our accuracy for both camera estimation and objects localization.

In the future, we plan to improve the object temporal correlation by fully exploiting the dense visual information. Currently, the camera and objects tracking are implemented successively in our system. We are also going to model them into a fully-integrated optimization framework such that the estimation for both camera and dynamic objects can benefit from each other.

\section{Acknowledgment}
This work was supported by the Hong Kong Research Grants Council Early Career Scheme under project no. 26201616. The authors also thank Xiaozhi Chen for providing the 3DOP \cite{chen20153d} results on the KITTI tracking dataset.

\bibliographystyle{splncs}
\bibliography{egbib}
\end{document}